\documentclass[10pt,twocolumn,letterpaper]{article}

\usepackage[pagenumbers]{cvpr} 

\usepackage{graphicx}
\usepackage{amsmath}
\usepackage{amssymb}
\usepackage{booktabs}
\usepackage{algorithm} 
\usepackage[noend]{algpseudocode}
\usepackage{xcolor}
\usepackage[inline]{enumitem}
\usepackage{multirow}
\usepackage{caption}
\usepackage{subcaption}
\usepackage{float}
\usepackage{lipsum}
\usepackage{makecell}

\usepackage[pagebackref,breaklinks,colorlinks=true,urlcolor=blue]{hyperref}

\usepackage[capitalize]{cleveref}
\crefname{section}{Sec.}{Secs.}
\Crefname{section}{Section}{Sections}
\Crefname{table}{Table}{Tables}
\crefname{table}{Tab.}{Tabs.}

\newcommand{\pstart}[1]{\textbf{#1.\enspace}}
\newcommand{\textvar}[1]{$l_{#1}$}
\newcommand{\imagevar}[1]{$I_{#1}$}

\def\cvprPaperID{11595} 
\def\confName{CVPR}
\def\confYear{2023}

\begin{document}

\title{Invariant Learning via Diffusion Dreamed Distribution Shifts}

\author{
  Priyatham Kattakinda\\
  University of Maryland \\
  \texttt{pkattaki@umd.edu}
  \and
  Alexander Levine\\
  University of Maryland \\
  \texttt{alevine0@cs.umd.edu}
  \and
  Soheil Feizi\\
  University of Maryland \\
  \texttt{sfeizi@cs.umd.edu}
}


\twocolumn[{%
\renewcommand\twocolumn[1][]{#1}%
\maketitle
\begin{center}
    \centering
    \captionsetup{type=figure}
    \includegraphics[width=\textwidth]{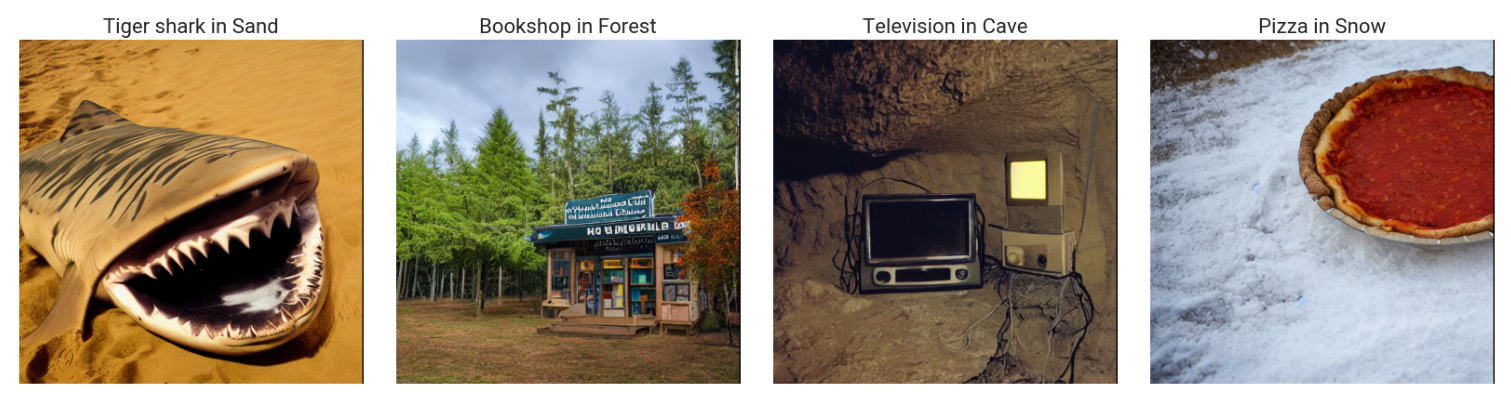}
    \captionof{figure}{Some images in our dataset that are difficult if not impossible to find in the real world.}
    \label{fig:d3s-surprising}
\end{center}%
}]

\begin{abstract}
Though the background is an important signal for image classification, over reliance on it can lead to incorrect predictions when spurious correlations between foreground and background are broken at test time. Training on a dataset where these correlations are unbiased would lead to more robust models. In this paper, we propose such a dataset called Diffusion Dreamed Distribution Shifts (D3S). D3S consists of synthetic images generated through StableDiffusion using text prompts and image guides obtained by pasting a sample foreground image onto a background template image. Using this scalable approach we generate 120K images of objects from all 1000 ImageNet classes in 10 diverse backgrounds. Due to the incredible photorealism of the diffusion model, our images are much closer to natural images than previous synthetic datasets. D3S contains a validation set of more than 17K images whose labels are human-verified in an MTurk study. Using the validation set, we evaluate several popular DNN image classifiers and find that the classification performance of models generally suffers on our background diverse images. Next, we leverage the foreground \& background labels in D3S to learn a foreground (background) representation that is invariant to changes in background (foreground) by penalizing the mutual information between the foreground (background) features and the background (foreground) labels. Linear classifiers trained on these features to predict foreground (background) from foreground (background) have high accuracies at 82.9\% (93.8\%), while classifiers that predict these labels from background and foreground have a much lower accuracy of 2.4\% and 45.6\% respectively. This suggests that our foreground and background features are well disentangled. We further test the efficacy of these representations by training classifiers on a task with strong spurious correlations.

\end{abstract}


\section{Introduction}
\label{sec:introduction}

Large, diverse datasets are crucial to training robust machine learning models. In the absence of diverse data, these models are prone to learning ``shortcuts'' that do no generalize well to real-world conditions \cite{geirhos2020shortcut}. This pathological behavior is particularly exacerbated when there are simple, yet incorrect decision rules that exploit spurious correlations in data \cite{beery2018recognition, sagawa2019distributionally, liu2015deep, jabbour2020xray, agarwal2020towards, zech2018variable}. A large body of work exists to detect these spurious correlations \cite{singla2021salient, wong2021leveraging, plumb2022finding, nushi2018towards,moayeri2022hard,moayeri2022comprehensive,kattakinda2022focus} and circumvent them \cite{kirichenko2022last, Zhang_2021_CVPR, abid2022meaningfully, asgari2022masktune}. In learning tasks where collecting rich, representative datasets is expensive, these methods play an important role in improving generalization performance of machine learning models. Nevertheless, if the data is diverse enough, one need not rely on such methods. Data augmentations such as random flipping~\cite{he2016deep}, MixUp~\cite{zhang2018mixup}, AugMix~\cite{hendrycks2020augmix}, Autoaugment~\cite{cubuk2019autoaugment}, AugMax~\cite{wang2021augmax} among many others are a relatively cheap way to improve sample diversity. However, these methods are limited in their ability to break spurious correlations between high-level semantic features such as those due to co-occurrence of two classes of objects or an object predominantly occurring in a particular environment. Consequently, the distribution shifts introduced by these augmentations fall far short of the demands of robustness in real-world settings.

Therefore, we leverage the phenomenal advances in photorealistic image generation using diffusion models. These models can generate a wide variety of images and offer a great degree of control over generation through text prompts and image guides. In our work, we construct a dataset called Diffusion Dreamed Distribution Shifts (D3S) consisting of synthetic images of objects in diverse backgrounds using StableDiffusion~\cite{rombach2022high}. Our dataset is as follows:
\begin{equation}
    {\left\{x^{(i)}, l^{(i)}_{bg}, l^{(i)}_{fg}\right\}}_{i=1}^{120000}
\end{equation}
where $x^{(i)}$ is a synthetic image generated through the diffusion model and $l_{fg}$, $l^{(i)}_{bg}$ are its foreground and background labels respectively. Note that the set of foreground labels $l^{(i)}_{fg}$ is same as that of ImageNet, so our dataset has 1000 foreground classes and the background labels $l^{(i)} \in $ \{on grass, on a road, in a forest, in water, in a cave, in sand, indoors, in snow, in rain, at night\} (10 background classes). 

Given a $l_{fg}$ and $l_{bg}$ drawn from the corresponding label sets, we generate an image of the object $l_{fg}$ as foreground in the background $l_{bg}$ as follows:
\begin{enumerate}
    \item Construct a text prompt of the form: ``\textit{a photo of a \{$l_{fg}$\}, \{definition($l_{fg}$)\}, \{$l_{bg}$\}}''.
    \item Generate an image guide by pasting an image of the foreground $l_{fg}$, drawn from ImageNet, onto a template image of the background ${l_{bg}}$.
    \item Use StableDiffusion to generate the desired image by conditioning it on the above text prompt as well as the guide image.
\end{enumerate}
Since we do the above procedure for every possible pair of foreground and background, the correlations between them in our dataset are controlled and unbiased. This is in contrast to most natural image datasets like ImageNet~\cite{deng2009imagenet}, MS COCO~\cite{lin2014microsoft} etc., which predominantly capture objects in environments/backgrounds that are likely to contain said objects. \Cref{fig:d3s-surprising} shows some samples from our dataset whose real world counterparts would be difficult if not impossible to find.

Admittedly, the images generated by the diffusion model are not always perfect. In particular, they may not contain the required object. This is unacceptable for evaluating models reliably. Therefore, we set aside a subset of 20K images for validation and further improve their correctness by conducting a study on Amazon Mechanical Turk. Human workers are asked to check that each of these images contain the correct object and if an image does not meet this standard, it is removed from the validation set. In the end, we are left with 17,866 images in the validation set with highly reliable labels.

We evaluate a range of popular pretrained models including CNNs, transformers, self-supervised, semi-supervised, semi-weakly supervised models, and, zero shot classifiers using CLIP encoders on our validation set. We find that all the models except the CLIP ones perform worse by $\sim$10-15\% compared to their accuracy on ImageNet. This suggests that these models are not robust to the foreground-background distribution shifts in our dataset.

We further leverage D3S to learn a feature extractor that extracts separate representations for foregrounds and backgrounds that are invariant with respect to the changes in the other. Concretely, given an image $X$, we want its foreground features $Z_{fg}$ and background features $Z_{fg}$ to be independent of the background label $L_{bg}$ and foreground label $L_{fg}$ respectively. This is done by enforcing the mutual information between the foreground (background) label and the background (foreground) feature to be 0: 
\begin{align}\label{eq:zero_mi-intro}
    I(Z_{fg};L_{bg}) = 0 && I(Z_{bg};L_{fg}) = 0
\end{align}
As we use Wasserstein distances as regularization terms to learn a feature extractor that (approximately) obeys the above constraints, we call our method Wasserstein disentanglement. \Cref{tab:dis-intro} shows the remarkable effectiveness of our disentanglement method. Predicting the foreground (background) with the corresponding foreground (background) invariant feature gives accuracy, while the other way around is ineffective.
\begin{table}[h]
    \centering
    \begin{tabular}{cccc}
         $A^{fg}_{fg}$ & $A^{fg}_{bg}$ & $A^{bg}_{bg}$ & $A^{bg}_{fg}$ \\[1mm]
         \hline
         82.9\% & 2.4\% & 93.8\%  & 45.6\%\\
    \end{tabular}
    \caption{Accuracy of predicting fg and bg labels from fg or bg features. $A_{i}^j$ refers to accuracy of predicting label $i$ from feature $j$}
    \label{tab:dis-intro}
\end{table}

We further test the efficacy of our invariant feature extractor in the presence of strong spurious correlations. Concretely, linear classifiers on top these features using images in which each foreground is exclusively present in a single unique background. Therefore, both the foreground and background can be predicted by looking at the other alone and are a ``shortcut'' for each other. We observe that for both foreground and background prediction, our invariant features are resilient to the artificially introduced spurious correlation and achieve an average accuracy of $95.88\%$ across all tasks while the baseline model pretrained on ImageNet only achieves $78.44\%$ as it fails to disregard the shortcuts. 

\section{Background}
\label{sec:background}

\subsection{Foreground-background distribution shifts}
\label{sec:background-distribution-shifts}
Several prior works have proposed ImageNet-like datasets with artificially controlled correlations between foreground and background features. However, these projects vary in the sophistication of the methods used to generate the images, and thus in the quality/naturalness of the images in the generated datasets. We include samples of each method in the appendix. As a simple method, Zhu \textit{et al.} \cite{zhu2017object} uses bounding boxes to construct separate datasets consisting of only foreground, and only background, objects from ImageNet. This allows for separate networks to be trained that attend to foreground or background features alone. Masked pixels are simply blacked out of the images.

Xiao \textit{et al.} \cite{xiao2021noise} constructs artificial images by overlaying the foreground from one image onto the background of another. They combine foregrounds and backgrounds of images with disparate class labels, for training and assessing models that attend mostly to foreground features, rather than (possible irrelevant) background features. The foregrounds are selected using GrabCut \cite{10.1145/1186562.1015720} along with bounding box annotations, and holes left in the backgrounds are filled using a tiling technique. However, because the foregrounds and backgrounds are combined using simple binary masks, it tends to produce unnatural-looking images.

Kattakinda and Feizi \cite{kattakinda2022focus} take a complementary approach in FOCUS, a dataset of systematically collected natural images from the Web which are curated to include uncommon backgrounds for a given foreground. This allows for the distribution of background environments to be controlled during evaluation. Foregrounds are labeled using 10 course-grained classes, but these can can each be mapped to sets of ImageNet classes, allowing for pretrained ImageNet classifiers to be assessed. Note that both \cite{zhu2017object} and \cite{xiao2021noise} also use such course-grained labels.
\subsection{Text-to-image diffusion models}
\label{sec:background-diffusion-models}
A diffusion probabilistic model \cite{sohl2015deep,ho2020denoising} is a type of generative model based on the idea of sequential denoising of images. 
For an image $x_0$, consider a sequence of noisy copies of the image $x_1$, ... $x_T$, defined as:
\begin{equation}
    x_{t} \sim  \mathcal{N}(\mu = \sqrt{1-\beta_t}x_{t-1}, \Sigma = \beta_t I  )
\end{equation}
This is known as the \textit{forward diffusion process}. For properly-chosen values of $\beta$, this means that $x_T$ is approximately distributed as $N(0,I)$. Then, by learning a \textit{reverse diffusion process} (parameterized by network parameters $\theta$):
\begin{equation}
    x_{t-1} \sim  \mathcal{N}(\mu = \mu_{\theta}(x_{t}, t), \Sigma = \Sigma_{\theta}(x_{t}, t)  )
\end{equation}
one can generate new samples $x_0$ from the training distribution, starting with generated Gaussian noise at $x_T$. 
This simple approach can be modified to allow for conditional sample generation \cite{sohl2015deep,dhariwal2021diffusion}, where samples are generated according to a distribution $P'(x_0) \propto P(x_0) \cdot r(x_0)$, where $P$ is the distribution modeled by the diffusion model, and $r$ represents some external guidance signal.  This can be used, for example along with CLIP~\cite{radford2021learning}, to generate photo realistic images matching text prompts~\cite{ramesh2021zero,ramesh2022hierarchical,saharia2022photorealistic,rombach2022high}.

Image diffusion models can also be conditioned on images by simply running forward diffusion for some limited number of steps $t < T$ on the prompt image, and then running reverse diffusion for $t$ steps. This produces an output that has some commonality with the input image, but is pushed towards the distribution modeled by the diffusion model as well as any conditioning. This method is used by diffusion-based image-editing methods, as discussed below.

\subsection{Image editing using diffusion models}
\label{sec:background-image-editing}
Several works have explored using diffusion probabilistic models to make photo-realistic edits to images. \cite{meng2021sdedit,DBLP:conf/icml/NicholDRSMMSC22,saharia2022palette,avrahami2022blended} SDEdit \cite{meng2021sdedit} allows for hand-drawn sketches to be added to photographs before the images are passed through forward diffusion; reverse diffusion then produces photorealistic images including the sketched objects. (Photographic components of the original image are masked to ensure that they are returned unmodified in the final images.) Nichol et al. \cite{DBLP:conf/icml/NicholDRSMMSC22} apply SDEdit to their GLIDE model, which allows for text-based conditioning on the image generation, using CLIP embeddings; Avrahami \etal \cite{avrahami2022blended} demonstrate a similar application of diffusion models. Saharia \etal \cite{saharia2022palette} explores using diffusion models for various image-to-image tasks such as inpainting and colorization. Ramesh \etal \cite{ramesh2022hierarchical} proposes interpolating CLIP embeddings of images as prompts to a diffusion model in order to interpolate between between the images. Recently, Ruiz~\etal \cite{ruiz2022dreambooth} has proposed DreamBooth, an approach for inserting objects from user-provided images into generated images via fine-tuning the diffusion model, while Gal \etal 
\cite{gal2022image} has proposed an approach that inverts a CLIP-conditioned diffusion model to find a ``pseudo-word'' word embedding that describes an object in a set of images; this pseudo-word can then be used to prompt new images containing the object. Note that both \cite{gal2022image, ruiz2022dreambooth} require 3-5 samples of the ``same'' object for targeted generation of images of that object in various environment. In contrast, our approach is much simpler and scalable as it does not involve any finetuning of the diffusion model.


\section{D3S: Diffusion Dreamed Distribution Shifts}
\label{sec:d3s}


\subsection{Generating diverse images using diffusion}
\label{sec:d3s-algorithm}

\begin{figure*}[t]
    \centering
    \includegraphics[scale=0.135]{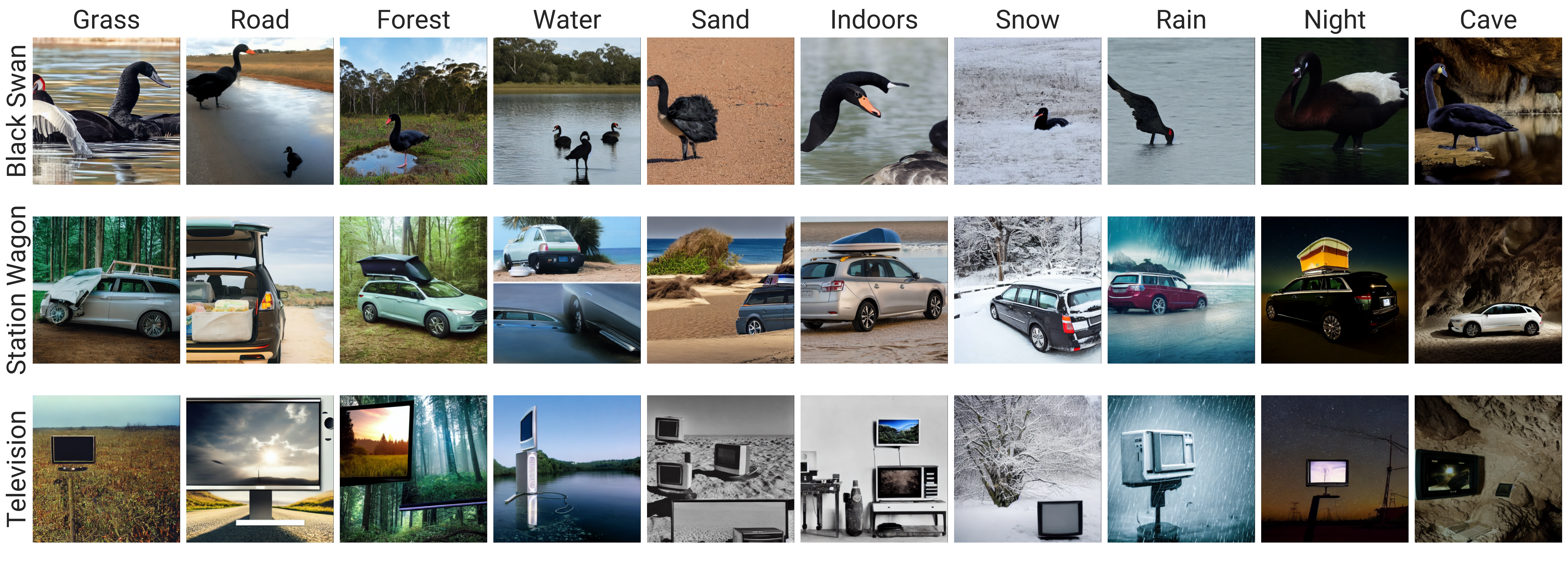}
    \caption{Our dataset D3S contains images of all 1000 classes of objects in ImageNet in various backgrounds. These images are generated synthetically by specific prompts and initialization images to StableDiffusion~\cite{rombach2022high}. See \cref{algorithm:d3s}, \cref{sec:d3s-algorithm} for details.}
    \label{fig:d3s-samples}
\end{figure*}

We use StableDiffusion \cite{rombach2022high} to generate our dataset Diffusion Dreamed Distribution Shifts (D3S) which consists of images of objects (foregrounds) in diverse environments (backgrounds). In this paper, we use all 1000 ImageNet classes as foregrounds and the set $\mathcal{B}$ = \{on grass, on a road, in a forest, in water, in a cave, in sand, indoors, in snow, in rain, at night\} as backgrounds. This list of backgrounds is inspired by FOCUS~\cite{kattakinda2022focus} and is not meant to be exhaustive. We believe that it is a good starting point and already achieves a great deal of diversity (see \cref{fig:d3s-samples}). As described in \cref{sec:background-diffusion-models}, the outputs of StableDiffusion are conditioned, optionally, on text prompts and image guides. We use both these modes of guidance in a scalable fashion (details below) to attain high label accuracy and background diversity.

\begin{figure}
    \centering
    \begin{subfigure}[b]{0.45\textwidth}
    \captionsetup{justification=centering}
    \includegraphics[trim={0 2cm 0 2cm}, clip, scale=0.45]{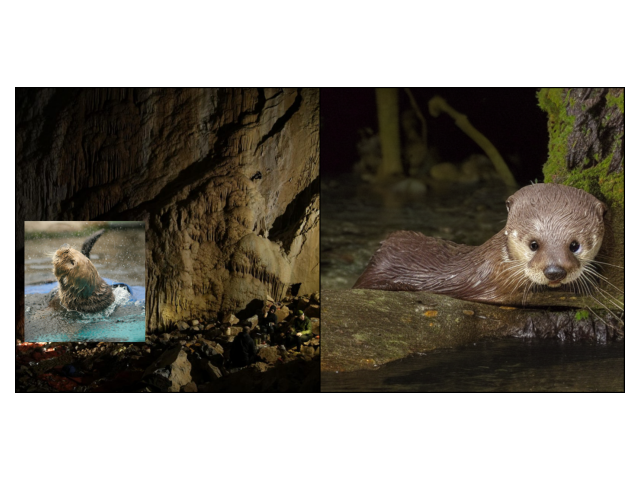}
    \caption{\textit{``a photo of a otter, freshwater carnivorous mammal having webbed and clawed feet and dark brown fur, in a cave''}}
    \label{fig:d3s-io-a}
    \end{subfigure}
    
    \begin{subfigure}[b]{0.45\textwidth}
    \captionsetup{justification=centering}
    \includegraphics[trim={0 2cm 0 2cm}, clip, scale=0.45]{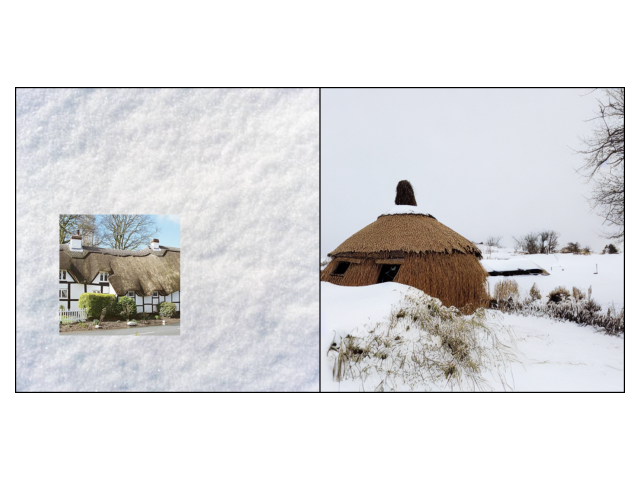}
    \caption{\textit{``a photo of a thatch, a house roof made with a plant material (as straw), in snow''}}
    \label{fig:d3s-io-b}
    \end{subfigure}
    \caption{Image guides and text prompts we use for generating D3S. In each subfigure, the guide image is on the left, the prompt we use is in the caption and the corresponding output image of the diffusion model is on the right.}
    \label{fig:d3s-io}
\end{figure}

\pstart{Image guides} We collect a small set ($B$) of background templates; three for each of the backgrounds in $\mathcal{B}$ (templates are show in the appendix). To generate an image with the foreground label \textvar{fg} and the background label \textvar{bg}, we first sample an image \imagevar{fg} from ImageNet that contains \textvar{fg} and a background image \imagevar{bg} corresponding to \textvar{bg} from $B$. Our guide image $G$ is constructed by simply downsampling \imagevar{fg} and pasting it on \imagevar{bg} at a random location. See \cref{fig:d3s-io} for some examples of our image guides. Yun \etal~\cite{yun2019cutmix} use a similar strategy in their data augmentation called CutMix. Our guide images are different from CutMix in two ways:  \begin{enumerate*}
    \item We paste the entirety of a downsampled \imagevar{fg} onto \imagevar{bg}, not just a portion of it
    \item We are not combining two samples in the base dataset (ImageNet in our case); \imagevar{bg} is sampled from a separate set of background templates.
\end{enumerate*}
The image $G$ acts as the guide for stochastic image editing (\cref{sec:background-diffusion-models}). Since we merely seek to produce an image of \textvar{fg} in \textvar{bg}, faithfulness to especially low-level attributes of $G$ is not relevant. On the other hand, we want the output to be realistic, so we choose a high noise strength ($t_0$) of 0.9. This is the variance of the noise added to the guide image and determines the time step at which the backward diffusion process in run. In practice, we observe that this leads to the diffusion model reproducing global attributes like color temperature of $G$ and the correct background in virtually all cases. We also tried MixUp~\cite{yao2022improving, zhang2018mixup} between \imagevar{fg} and \imagevar{bg}, but we find the results to be inferior. We postulate that this is because the image encoder in StableDiffusion sees a significant distribution shift, even at a per-pixel level, and hence fails to meaningfully encode the mixed up image.

\pstart{Text prompts}Prompt engineering \cite{liu2022design, zhou2022learning} has attracted considerable attention with the rise in popularity of text conditional generative models. To meet our needs of simplicity and flexibility, we use the following template for our text prompts:

{
    \centering
    \textit{a photo of a \{$l_{fg}$\}, \{definition($l_{fg}$)\}, \{$l_{bg}$\}}
    \par
}

\noindent \textvar{fg} is sampled from the 1000 class labels in ImageNet and \textvar{bg} is sampled from $\mathcal{B}$. Given that images are seldom captioned with just a single word, Radford \etal~\cite{radford2021learning} recommend using \textit{``a photo of a \{$l_{fg}$\}''} rather than a single word like ``\textvar{fg}'' for prompting models trained on large scale vision-natural language datasets. In addition, we find that using the word ``photo'' as opposed to other, alternatives like ``image'' or ``figure'' generates photo realistic outputs. To disambiguate polysemic labels we also include the definition of the class from WordNet~\cite{miller1995wordnet}. Lastly, all our prompts end with the suffix \textvar{bg} to encourage StableDiffusion to place the object in a particular background. See the captions in \cref{fig:d3s-io-a,fig:d3s-io-b} for examples of our text prompts.

See \cref{algorithm:d3s} for an outline of our procedure to generate D3S. This approach is extremely scalable; we generate 120000 images on 8 NVIDIA RTX A4000 GPUs in under a day. \Cref{fig:d3s-samples} shows some samples from our dataset.


\pstart{Sample Diversity} ImageNet is a diverse dataset with objects in various poses, illumination, locations etc. But, its strength of having all natural images can be a detriment. For one, it is subject to the biases of its source: the internet. Luccioni and Rolnick~\cite{luccioni2022bugs} demonstrate serious geographical and cultural biases in the biodiversity of images in ImageNet. For example, as ImageNet was collected by scraping Flickr, many images of fish are of people holding a dead fish. In addition, as discussed in Kattakinda and Feizi~\cite{kattakinda2022focus}, one must be mindful of the sampling procedure used to collect these datasets or risk not capturing particularly uncommon scenarios. Our images, though synthetic, are near photorealistic, thanks to the capabilities of StableDiffusion and augment datasets like ImageNet by expanding diversity in a targeted fashion along one specific dimension: foreground vs. background correlations. That said, we provide some objective evidence for overall diversity of D3S images by computing LPIPS distances~\cite{zhang2018unreasonable} between different pairs of images of the same class. When both images are drawn from ImageNet, the mean distance is 0.706, when they are both drawn from our dataset D3S it is 0.716 and when one image each is drawn from the two datasets, it is 0.730. This shows that the images in our dataset are, perceptually, at least as diverse as those in ImageNet. See supplementary material for a histogram of these LPIPS distances.


\subsection{Creating the D3S benchmark}
\label{sec:d3s-benchmark}

The images generated through \cref{algorithm:d3s} have some label noise because StableDiffusion does not always place the correct foreground in the output. To create an accurate benchmark, we separate out a subset of 20000 images from D3S. We conduct an Amazon Mechanical Turk study on this subset, where we show human workers both the image guides and the corresponding outputs. Workers are asked to tag the images where the foreground in the image guide is not reproduced in the output. In addition, workers are also tasked to filter out inappropriate material. At the end of the study, we are left with 17,866 of the 20000 we started with. These images constitute the D3S benchmark and we use it for evaluating models throughout this paper. See supplementary material for more details about this AMT study.

\begin{algorithm}[h]
\caption{Producing images with diverse backgrounds}
\label{algorithm:d3s}
\hspace*{\algorithmicindent} \textbf{Input} ImageNet $D$, Background templates $B$, Number of diverse samples $n$, Diffusion Model $M$  \\
\hspace*{\algorithmicindent} \textbf{Output} Diverse dataset $\tilde{D}$
\begin{algorithmic}[1]
\Procedure{BGShift}{$D, B, n$}
\While {\text{ len($\tilde{D}$) != n}}
\State $I_{fg}, l_{fg}$ $\gets$ \text{Sample a foreground image from $D$}
\State $I_{bg}, l_{bg}$ $\gets$ \text{Sample a background image from $B$}
\State $ I_{\text{guide}}\gets$ \text{Downsample and paste $I_{fg}$ on $I_{bg}$}
\State prompt $\gets$ \textit{``a photo of a \{$l_{fg}$\}, \{definition($l_{fg}$)\}, in \{$l_{bg}$\}''}
\State output $\gets M$(prompt, $I_{\text{guide}}$)
\State Add output to $\tilde{D}$
\EndWhile
\EndProcedure
\end{algorithmic}
\end{algorithm}

\begin{figure}
    \centering
    \includegraphics[scale=0.52]{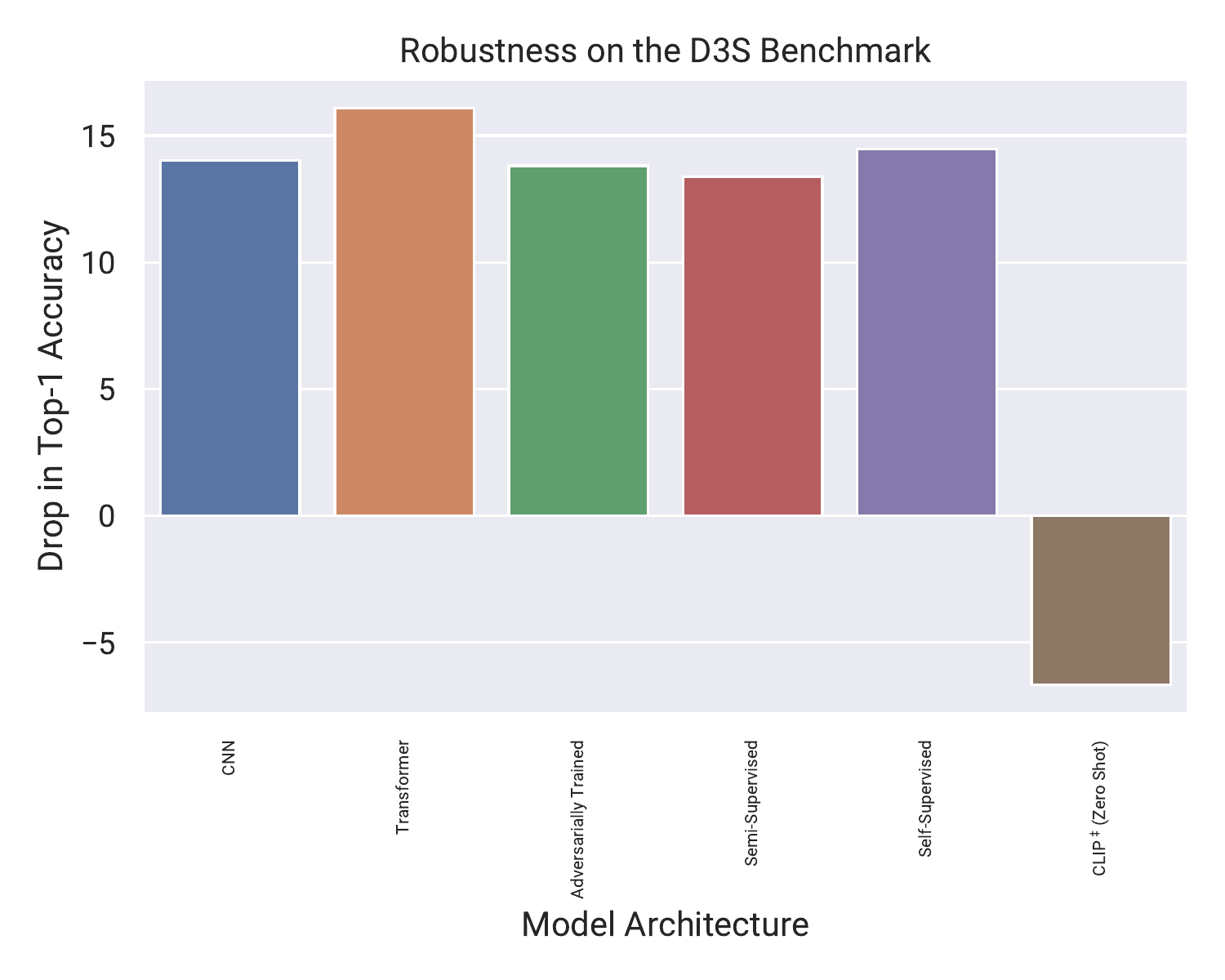}
    \caption{\textbf{Most models have poor generalization to the diverse images in our benchmark}. Figure shows the difference in top-1 accuracy on ImageNet and our benchmark for various models trained on ImageNet (except the CLIP models; see text for details).} 
    \label{fig:all-gaps}
\end{figure}


\section{Evaluating pretrained models on D3S}
\label{sec:experiments-d3s}


We evaluate the following image classifiers on our D3S benchmark:
\begin{enumerate*}[label=(\roman*)]
\item Supervised CNNs (ResNet~\cite{he2016deep}, EfficientNet~\cite{tan2019efficientnet}, Inception-ResNet-v2~\cite{szegedy2017inception}, MobileNet~\cite{howard2018inverted, howard2019searching}, DenseNet~\cite{huang2017densely}, ResNeXt~\cite{xie2017aggregated}, ResNetv2~\cite{he2016identity})
\item Supervised vision transformers (ViT~\cite{dosovitskiy2021an}, DeiT~\cite{touvron2021training})
\item Adversarially robust models (Adversarial Inception networks \cite{kurakin2018adversarial}, Robust ResNets\cite{robustness})
\item Semi-supervised \& Semi-weakly supervised models (SSL, SWSL ResNets~\cite{zeki2019billion})
\item Self-supervised models (BEiT~\cite{bao2022beit}, DINO~\cite{caron2021emerging})
\item Clip zero shot classifiers ~\cite{radford2021learning}.
\end{enumerate*} \Cref{fig:gaps} shows the difference in top-1 accuracy of these models on the validation set of ImageNet and on our D3S benchmark, averaged per architecture type. 
The drop in accuracy is significant at $\sim$ 10-15\% for all model types except the CLIP ones. We postulate that this is because StableDiffusion uses the same text tokenizer as CLIP and the prompts used for zero shot classfication are similar to the prompts we use in generating images in D3S; both have the prefix ``a photo of a \{fg\}''. The full chart with accuracy drops for individual models and details about the backgrounds that we identified as the most and least predictive of the foregrounds for all these models are included in the supplementary material.

\section{Breaking correlations between foregrounds and backgrounds}
\label{sec:experiments-disentanglement}

Beery \etal~\cite{beery2018recognition} demonstrate ``shortcut learning'' in classification and object detection models using the Caltech Camera Traps dataset. This dataset has images of animals taken from camera traps in fixed locations. Since the cameras are fixed, all the images from a single trap have the same background. Furthermore, a given camera might capture a particular animal species far more often than others owing to the prevalence of that species in that location. For example, a camera trap in a desert might capture many coyotes, while a camera on a beach might seldom capture cows. They observe that DNNs perform poorly on ``trans-locations''; locations not seen during training. This behavior can be explained by observing that location is a shortcut for species classification and as such is not robust to new, unseen locations. Therefore, breaking correlations between foregrounds and backgrounds is crucial for robust models. In this section, we describe how we achieve just that using our dataset D3S which has diverse images with both foreground and background labels.

Denote by $(X, L_{fg}, L_{bg})$ an image, its foreground label and its background label, respectively. Our goal is to learn a neural network $h_{\theta}$ (with parameters $\theta$) that computes a disentangled feature representation of $X$ as $(Z_{fg}, Z_{bg})$ such that: (i) we can predict $L_{fg}$ from $Z_{fg}$ and $L_{bg}$ from $Z_{bg}$, respectively, and (ii) $Z_{fg}$ (or, $Z_{bg}$) contains no information about the label $L_{bg}$ (or, $L_{fg}$). To achieve the first goal, we use two linear classifiers $w_{fg}$ and $w_{bg}$ that are trained (along with $h_{\theta}$) using cross entropy loss. Our second goal can be written as:
\begin{align}\label{eq:zero_mi}
    I(Z_{fg};L_{bg}) = 0 && I(Z_{bg};L_{fg}) = 0
\end{align}
where $I(Y_1;Y_2)$ denotes the mutual information between two random variables $Y_1, Y_2$. $I(Y_1;Y_2) = 0 $ if and only if they are independent i.e., the joint distribution of $Y_1, Y_2$ (denoted by $\mathbb{P}_{Y_1Y_2}$) is equal to the product of their marginal distributions (denoted by $\mathbb{P}_{Y_1}\mathbb{P}_{Y_2}$). Therefore, \cref{eq:zero_mi} can be written as:
\begin{align}
    \mathbb{P}_{Z_{fg}L_{bg}} = \mathbb{P}_{Z_{fg}}\mathbb{P}_{L_{bg}} && \mathbb{P}_{Z_{bg}L_{fg}} = \mathbb{P}_{Z_{bg}}\mathbb{P}_{L_{fg}}
\end{align}
These constraints can be enforced by requiring that the Wasserstein distances ($W(.,.)$) between the distributions on both sides of the above equations be zero:
\begin{subequations}
\begin{align}
    W(\mathbb{P}_{Z_{fg}L_{bg}}, \mathbb{P}_{Z_{fg}}\mathbb{P}_{L_{bg}}) &= 0\label{eq:wasserstein1} \\
    W(\mathbb{P}_{Z_{bg}L_{fg}}, \mathbb{P}_{Z_{bg}}\mathbb{P}_{L_{fg}}) &= 0\label{eq:wasserstein2}
\end{align}
\end{subequations}
Following the work of Arjovsky \etal~\cite{arjovsky2017wasserstein}, we use the Kantorovich-Rubenstein dual form of the Wasserstein distance which allows us to write \cref{eq:wasserstein1,eq:wasserstein2} as:
\begin{subequations}
\begin{align}
    \max_{\|D_{fg}\|_L \leq 1}& \ \mathbb{E}_{z_{fg}, l_{bg} \sim \mathbb{P}_{Z_{fg}L_{bg} }}\left[D_{fg}(z_{fg}, l_{bg})\right] \nonumber \\ 
    &- \mathbb{E}_{z_{fg}, l_{bg} \sim \mathbb{P}_{Z_{fg}}\mathbb{P}_{L_{bg} }}\left[D_{fg}(z_{fg}, l_{bg})\right] = 0 \label{eq:wasserstein-dual1}\\
    \max_{\|D_{bg}\|_L \leq 1}& \ \mathbb{E}_{z_{bg}, l_{fg} \sim \mathbb{P}_{Z_{bg}L_{fg} }}\left[D_{bg}(z_{bg}, l_{fg})\right] \nonumber \\
    &- \mathbb{E}_{z_{bg}, l_{fg} \sim \mathbb{P}_{Z_{bg}}\mathbb{P}_{L_{fg} }}\left[D_{bg}(z_{bg}, l_{fg})\right] = 0 \label{eq:wasserstein-dual2}
\end{align}
\end{subequations}
where $D_{fg}$ is the critic for disentangling foreground features from background labels and, similarly, $D_{bg}$ is the critic for disentangling background features from foreground labels. Constraining the weights of the critics $D_{fg}$ and $D_{bg}$ to lie in a compact space would satisfy the norm bounds in in \cref{eq:wasserstein-dual1,eq:wasserstein-dual2}. Therefore, Arjovsky \etal~\cite{arjovsky2017wasserstein} propose weight clipping as a simple way to enforce the constraints. However, Gulrajani \etal~\cite{gulrajani2017improved} argue that this leads to issues such as capacity underuse and exploding/vanishing gradients in the critics. Instead they show that, with probability 1, an optimal critic has gradients of unit norm on any straight line between two points each drawn from the distributions between which the Wasserstein distance is being computed. They propose enforcing this by directly penalizing the norm of the gradient to be 1 using a squared loss and demonstrate that this is a better approach to ensure Lipschitzness of the critics than simple weight clipping. As such, we use the same gradient penalty in our work and this gives the following objective for critic $D_{fg}$:
\begin{multline}
    \min_{D_{fg}} \ \mathbb{E}_{z_{fg}, l_{bg} \sim \mathbb{P}_{Z_{fg}}\mathbb{P}_{L_{bg} }}\left[D_{fg}(z_{fg}, l_{bg})\right] \\
    - \mathbb{E}_{z_{fg}, l_{bg} \sim \mathbb{P}_{Z_{fg}L_{bg} }}\left[D_{fg}(z_{fg}, l_{bg})\right] \\ + \lambda \left(\|\nabla D_{fg} (\tilde{z}_{fg}, \tilde{l}_{bg})\|_2 -1 \right)^2 \label{eq:loss_d}
\end{multline}
where $(\tilde{z}_{fg}, \tilde{l}_{bg}) = \left(\epsilon z^{(1)}_{fg} + (1 - \epsilon) z^{(2)}_{fg}, \epsilon l^{(1)}_{bg} + (1 - \epsilon)l^{(2)}_{bg}\right)$ in which $\epsilon \sim U(0, 1), \left(z^{(1)}_{fg}, l^{(1)}_{bg}\right) \sim \mathbb{P}_{Z_{fg}}\mathbb{P}_{L_{bg}}$ and $\left(z^{(2)}_{fg}, l^{(2)}_{bg}\right) \sim \mathbb{P}_{Z_{fg}L_{bg} }$. The other critic $D_{bg}$ is also trained with a similar objective. Finally, our loss function for the feature extractor $h_{\theta}$ and the linear classifiers $w_{fg}, w_{bg}$ is:
\begin{align}
\min_{\theta, w_{fg}, w_{bg}} \ \mathbb{E}_{z_{fg}, l_{bg} \sim \mathbb{P}_{Z_{fg}L_{bg} }} \left[D_{fg}(z_{fg}, l_{bg}) \right] \nonumber \\
     -  \mathbb{E}_{z_{fg}, l_{bg} \sim \mathbb{P}_{Z_{fg}}\mathbb{P}_{L_{bg} }}\left[D_{fg}(z_{fg}, l_{bg}) \right. \label{eq:loss_h}\\
     \left. - \alpha \mathcal{L}_{ce}(w^T_{fg}z_{fg}, l_{fg}) - \alpha \mathcal{L}_{ce}(w^T_{bg}z_{bg}, l_{bg})\right] \nonumber
\end{align} \label{eq:total_loss}
where $\mathcal{L}_{ce}$ denotes cross entropy loss and $\alpha$ controls the relative importance of cross entropy loss and the Wasserstein distance loss. Note that this loss depends implicitly on $h_{\theta}$ as $(z_{fg}, z_{bg}) = h_{\theta}(x)$.

We call our method \textit{Wasserstein Disentanglement.} Shen \etal~\cite{shen2018wasserstein} use a similar objective to tackle the problem of domain adaptation. Our methods differ, in that whereas they use Wasserstein distance to learn features that are invariant across domains, we use it as a regularization term to minimize a proxy of mutual information (called the ``Wasserstein Dependency Measure''~\cite{ozair2019wasserstein}) between a set of features and labels.
However, we can compare the methods directly if we consider, for example, the foreground features as the features of interest, and the  background labels as the  ``domains.'' Shen \etal applies their method only in the case of binary domains (e.g., only two possible background labels).
Concretely, their loss function is equivalent to: $W(\mathbb{P}_{Z_{fg}(l_{bg} =1)}, \mathbb{P}_{Z_{fg}(l_{bg} =2)})$ where $1,2$ represent the two different background labels. Because this loss only considers two distributions of $Z_{fg}$, Shen \etal therefore uses a discriminator which is not conditioned on the domain label. By contrast, by conditioning our discriminator on labels, we can handle the multiclass case.


\begin{table}[hbtp]
    \centering
    \begin{tabular}{cccc}
    \hline
         Input & Predicted & \multirow{2}*{Dataset} & \multirow{2}*{Accuracy$^{\dagger}$}  \\
         Features & Attribute \\
         \hline
        $z_{fg}$ & $l_{fg}$ & ImageNet & 65\% (86.2\%) ($\uparrow$) \\
        $z_{fg}$ & $l_{fg}$ & D3S & 82.9\% (97.1\%) ($\uparrow$) \\
        $z_{bg}$ & $l_{fg}$ & ImageNet & 2.1\% (4.9\%) ($\downarrow$) \\
        $z_{bg}$ & $l_{fg}$ & D3S & 2.4\% (5.5\%) ($\downarrow$) \\
        $z_{bg}$ & $l_{bg}$ & D3S & 93.8\% ($\uparrow$) \\
        $z_{fg}$ & $l_{bg}$ & D3S & 45.6\% ($\downarrow$) \\
    \end{tabular}
    \caption{Classification accuracy of different linear classifiers trained that predict either foreground or background label using our learned partitioned foreground and background features. The arrows indicate desired behaviour: $\uparrow$ -- higher is better, $\downarrow$ -- lower is better. ($\dagger$ -- The numbers in brackets are top-5 accuracies) }
    \label{tab:accs_disentanglement}
\end{table}


\begin{table*}[ht!]
    \centering
    \addtolength{\tabcolsep}{-0.2em}
    \begin{tabular}{|c|c|c|c|c|c|c|}
    \hline
         \multirow{2}*{Model}  & Correlated $fg$ &  Anti-correlated $fg$& Correlated $bg$&  Anti-correlated $bg$& ImageNet &\multirow{2}*{Average}  \\
         &D3S Accuracy&D3S Accuracy & D3S Accuracy &D3S Accuracy & Accuracy &   \\
         \hline
        \textbf{Wass. Dis. - Correct} & 99.57\%& 97.4\%& 99.29\%& 90.36\%& 92.77\%&\textbf{95.88\%}\\
         \hline
       Wass. Dis. - All &99.86\%&18.39\%&99.86\%&78.56\%&62.84\%&71.90\%\\

         \hline
       Baseline&99.72\%&91.71\%&99.72\%&6.485\%&94.55\%&78.44\%\\
       \hline
    \end{tabular}
    \caption{Classification accuracy of linear classifiers trained that predict \textit{correlated} foreground or background label using our learned partitioned foreground and background features. Note that all foreground and ImageNet accuracies refer to ImageNet-9 classes. ImageNet accuracy refers to the accuracy of the linear head trained to predict D3S foreground labels when tested directly on ImageNet-9 validation images. ``Average''is the average of the 5 experiments (not weighted to dataset size). See text of Section \ref{sec:corr_exps} for a description of each model. }
    \label{tab:accs_corr}
\end{table*}

In our experiments, the feature extractor $h_{\theta}$ is a ResNet50~\cite{he2016deep} without its fully connected layer. We initialize this model with the ImageNet pretrained weights provided by PyTorch~\cite{paszke2019pytorch}. The output of this feature extractor is a 2048 dimensional vector which we partition into a 1798 dimensional vector for foreground features $x_{fg}$ and the remaining 250 dimensions for background features $x_{bg}$. Our discriminators are multilayer perceptrons with LeakyReLU activations~\cite{xu2015empirical}. We sample batches both from ImageNet and our dataset D3S. However, since ImageNet does not have background labels, we do not compute loss terms involving background labels for ImageNet images. Observe that computing the loss functions in \cref{eq:loss_d,eq:loss_h} involves sampling from 4 different distributions: $\mathbb{P}_{Z_{fg}L_{bg}}, \mathbb{P}_{Z_{bg}L_{fg}},  \mathbb{P}_{Z_{fg}}\mathbb{P}_{L_{bg}}, \mathbb{P}_{Z_{bg}}\mathbb{P}_{L_{fg} }$. Samples from the first two distributions can be obtained simply sampling a batch $\{x^{(i)}, l^{(i)}_{fg}, l^{(i)}_{bg}\}$ of images, their foreground and background labels respectively and, computing their features as $\left(z^{(i)}_{fg}, z^{(i)}_{bg}\right) = h_{\theta}(x^{(i)})$. For samples from the product distributions $\mathbb{P}_{Z_{fg}}\mathbb{P}_{L_{bg}}, \mathbb{P}_{Z_{bg}}\mathbb{P}_{L_{fg}}$, extracting features (using $h_{\theta}$) from these batches would give samples from the true joint distributions $\mathbb{P}_{X_{fg}L_{bg}}$ and $\mathbb{P}_{X_{bg}L_{fg}}$. For sampling from the product marginal distributions $\mathbb{P}_{X_{fg}}\mathbb{P}_{L_{bg}}$ and $\mathbb{P}_{X_{bg}}\mathbb{P}_{L_{fg}}$, we simply shuffle the labels in a batch. This causes the labels to be independent of the features while still being drawn from the correct marginal distributions $\mathbb{P}_{L_{fg}}$ and $\mathbb{P}_{L_{bg}}$. These samples allow us to compute unbiased estimates of the gradients of the loss functions in \cref{eq:loss_h,eq:loss_d}, which we use to update the parameters through gradient descent. In our training, we alternate between updating the critics $D_{fg}, D_{bg}$ and updating the feature extractor $h_{\theta}$ and the linear heads $w_{fg}, w_{bg}$. We train with Adam optimizer~\cite{kingma2015adam} with $\beta_1=0, \beta_2=0.9$ for 160K iterations with a batch size of 64. 

We then fix the feature extractor $h_{\theta}$ and train 4 linear classifiers, one for each of the following tasks: \begin{enumerate*}[label=(\roman*)]\item predicting $l_{fg}$ from $z_{fg}$ \item predicting $l_{bg}$ from  $z_{bg}$ \item $l_{fg}$ from $z_{bg}$ \item $l_{bg}$ from $z_{fg}$ \end{enumerate*}. Here, foreground prediction is a 1000-way classification problem while background prediction is 10-way. \Cref{tab:accs_disentanglement} shows the accuracy of these classifiers on both ImageNet and D3S. Clearly, the first two classifiers that predict the labels $l_{fg}, l_{bg}$ using $z_{fg}, z_{bg}$ respectively perform much better than their counterparts that predict these labels using $z_{bg}, z_{fg}$ (note that the order is flipped) respectively. This indicates that our feature extractor has learned to partition information about foreground and background into separate dimensions of the feature vector. One can use this partitioned feature vector to learn, for example, foreground classifiers that are robust to spurious background correlations as we show in the next section.

\section{Testing robustness to spurious correlations} \label{sec:corr_exps}
In order to measure the effectiveness of our disentanglement method on removing spurious correlations, we create a subset of D3S, in which background and foreground labels are perfectly correlated. For this experiment, we use course-grained IN-9 foreground labels \cite{xiao2021noise}, which organise a subset of the 1000 ImageNet classes into 9 super-classes. We then arbitrarily associate each IN-9 class with a single D3S background class (the specific permutation used is listed in the appendix). The resulting dataset, $D3S_{corr}$ consists of only D3S images with these nine pairs of foreground-background labels.
We also construct $D3S_{anticorr}$, which only consists of D3S images that are absent in $D3S_{corr}$.

We wish to investigate whether using the features of our Wasserstein Disentanglement model will allow us to generalize from $D3S_{corr}$ to $D3S_{anticorr}$, despite the spurious foreground-background correlations artificially introduced in $D3S_{corr}$. To test this, we trained linear models on the training set of $D3S_{corr}$ in the following configurations:
\begin{itemize}
\item Learning foreground (background) labels from the foreground (background) features of the Wassserstein Disentanglement model; this is the intended use of our  model (labeled as ``Wass. Dis. - Correct'' in Table \ref{tab:accs_corr}).
\item Learning both foreground and background labels from all the features of the Wassserstein Disentanglement model. This acts as an ablation study for the disentanglement (Labeled as ``Wass. Dis. - All'' in Table \ref{tab:accs_corr}). 
\item Learning both foreground and background labels from the penultimate-layer features of a pre-trained ImageNet model. (Labeled as ``Baseline'' in Table \ref{tab:accs_corr}).
\end{itemize}
\Cref{tab:accs_corr} shows the results of evaluating all of these models on the validation sets of $D3S_{corr}$ and $D3S_{anticorr}$. We find that training foreground labels on foreground features and vice-versa allows us to maintain high ($>90$\%) accuracy even when the correlation is broken at test time, showing the usefulness of the disentanglement. In contrast, training with all features resulted in catastrophically bad performance ($<$19\%) when predicting anti-correlated foregrounds, while the  baseline model had catastrophically bad performance when predicting anti-correlated backgrounds. This suggests that the model trained with all features was mostly relying on background information alone during linear-head training, while the baseline model was mostly relying on foreground information.
We note that generating D3S ultimately relies on the training dataset of the Stable Diffusion model, which is far larger than the ImageNet training set and likely contains a wider variety of combinations of foreground and background features. Our goal is not to compare the overall performance of models trained on D3S with models trained on ImageNet, but rather to show that one can use D3S, along with our Wasserstein Disentanglement method, to obtain independent foreground and background representations that can generalize across datasets.

We also evaluated D3S-trained models on the ``Backgrounds Challenge'' dataset from Xiao \etal \cite{xiao2021noise}, but observed somewhat poor performance on those artifical foreground-background pairs. This experiment is presented in the appendix. We hypothesise that the poor performance is due to the unnatural ``cut-and-paste'' nature of the images from Xiao \etal, discussed in Section \ref{sec:background-distribution-shifts}.


\section{Conclusion}
\label{sec:conclusion}

We have proposed a new dataset, Diffusion Dreamed Distribution Shifts (D3S), of 120K synthetic images with 1000 different objects (as in ImageNet) in 10 diverse backgrounds. We evaluate several image classification models on the validation subset of D3S in which the labels are verified by human workers. We find that these models are not robust to the foreground vs. background distribution shifts in our dataset. We have also proposed a method to learn invariant features for foreground and background by penalizing the mutual information between the features and the background and foreground labels respectively. We demonstrate the effectiveness of these features through an experiment where we the classification task is imbued with strong spurious correlations and is extremely susceptible to shortcut learning. Our feature extractor successfully ignores these correlations and a linear classifier trained on top these features is able to achieve high classification accuracy even when said correlations are absent in the test set.

\section{Acknowledgements}
This project was supported in part by NSF CAREER AWARD 1942230, HR001119S0026 (GARD), ONR YIP award N00014-22-1-2271, Army Grant No. W911NF2120076, the NSF award CCF2212458, Meta grant 23010098. The authors would like to thank Neha Kalibhat, Gaurang Sriramanan, Wenxiao Wang for helpful discussions.

{\small
\bibliographystyle{ieee_fullname}
\bibliography{mainbib}
}
\pagebreak
\appendix

\onecolumn
\section{More samples from D3S}

\begin{figure*}[h]
    \centering
    \includegraphics[width=0.9\linewidth]{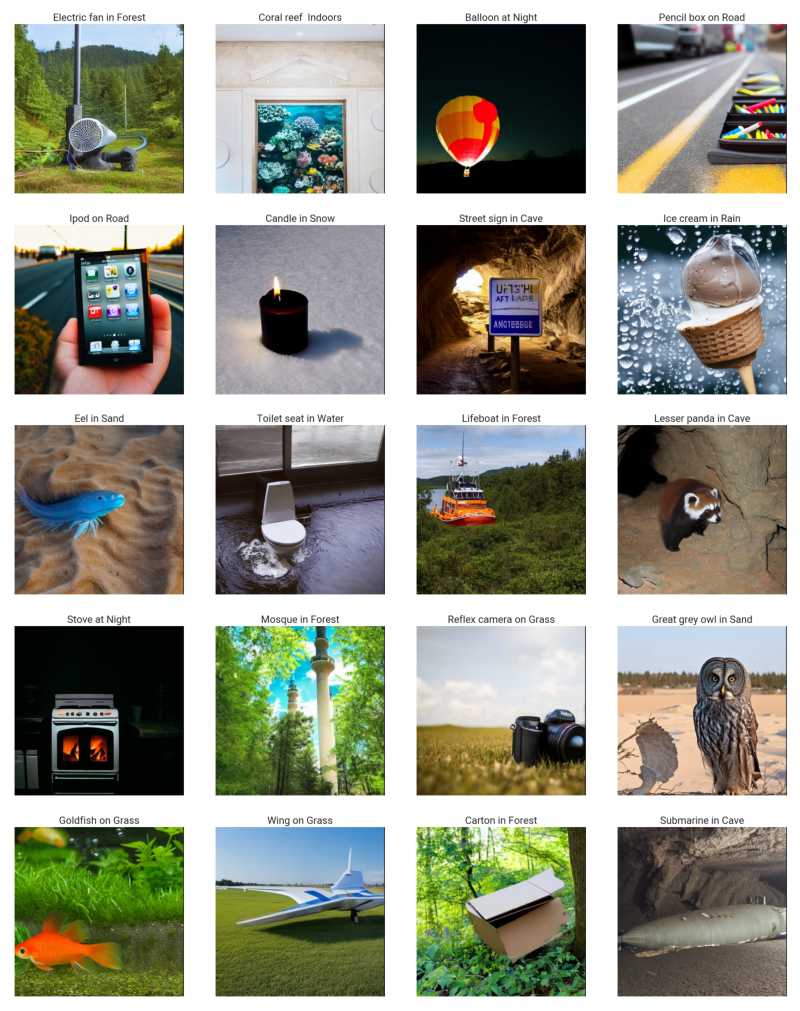}
    \caption{Some more samples from D3S that are difficult to find in the real world.}
    \label{fig:more-surprising}
\end{figure*}
\pagebreak
\noindent \textbf{Sample quality. } Some observations and issues with the samples in D3S:
\begin{enumerate}
    \item Almost all the images in D3S feature the intended foreground object prominently and no other objects . This means that our images do not have the problem of multiple conflicting foregrounds.
    \item Some objects are not true to their real life counterparts. This includes images where the objects are not of the correct shape, have missing parts etc., See the images of the wing, fan above.
    \item StableDiffusion struggles to generate coherent text as can be seen in the street sign image.
    \item Occasionally, the foreground or background produced in the output of the diffusion model is not correct. We observe that the noise in foreground labels far exceeds that in background noise. Since we are primarily interested in foreground classification, we perform an MTurk study to remove the foreground noise in the validation split of our dataset.
    \item Objects that require regular repeating patterns often have weird artifacts as in the fan image above and in the image of the fence in the screenshot below.
\end{enumerate}

\section{Amazon Mechanical Turk study}
\begin{figure*}[h]
\centering
\includegraphics[width=0.9\linewidth]{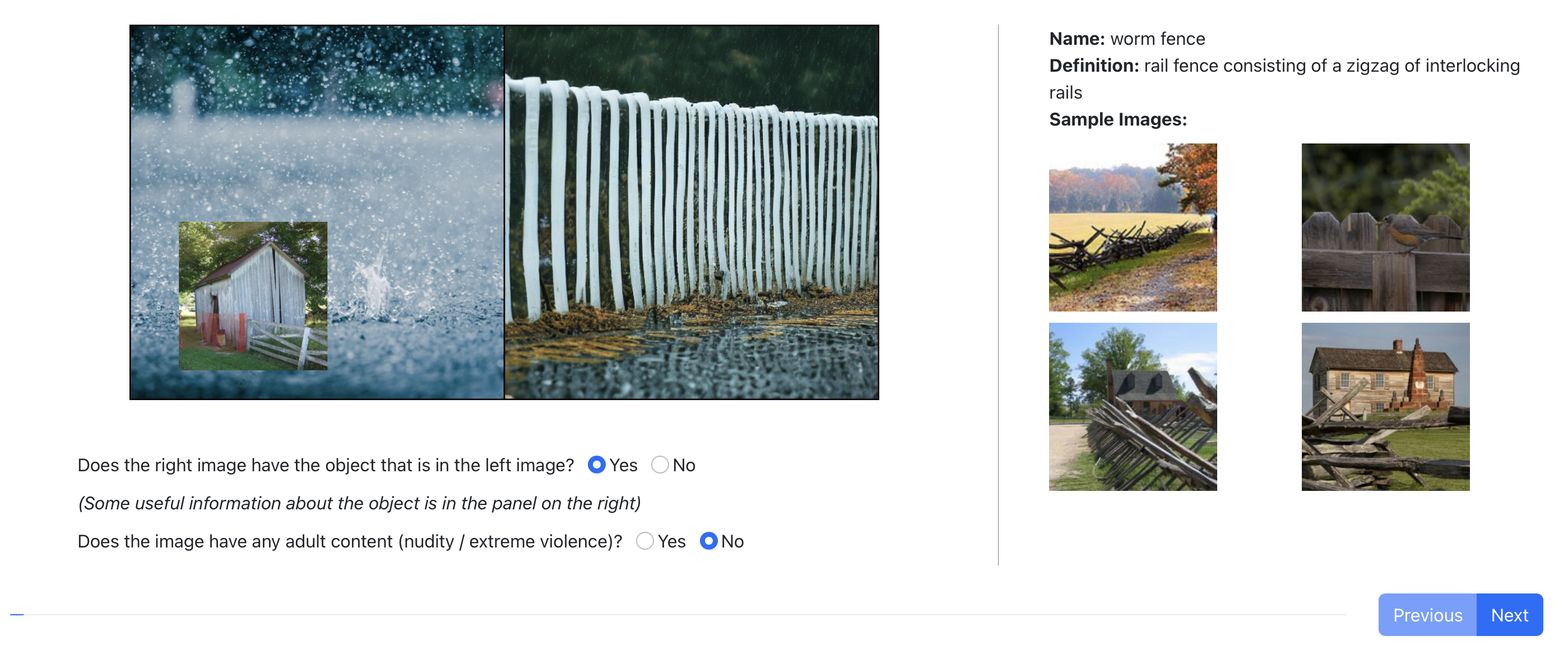}
\caption{UI used by human workers on Amazon Mechanical Turk to identify images with inaccurate foregrounds or NSFW material.}
\label{fig:mturk-ui}
\end{figure*}

The outputs from StableDiffusion~\cite{rombach2022high} are some times inaccurate; they may not have the intended object. To create a high quality validation split in our dataset D3S, we first set aside a subset of 20000 images. In this set, we identify images with incorrect foregrounds or NSFW material through a study on Amazon Mechanical Turk. Human workers are asked two binary questions: one for accuracy of foreground object and one for the presence of NSFW material. \Cref{fig:mturk-ui} shows a screenshot of the user interface shown to the workers in our study. Note that we also show the WordNet~\cite{miller1995wordnet} definition of the foreground class, and four sample images from ImageNet~\cite{deng2009imagenet} to aid workers in their assessment. 

After the study, we find that 10.2\% of the images have incorrect foreground and around 1\% of the images have been flagged as containing NSFW material. We filter out these images and call the remaining 17866 images the ``D3S benchmark''. All the model evaluations in our paper are on this subset.
\pagebreak
\section{Background templates}

\begin{figure*}[h]
    \centering
    
    \begin{subfigure}[h]{0.45\textwidth}
    \includegraphics[trim={0 26cm 0 0}, clip, scale=0.5]{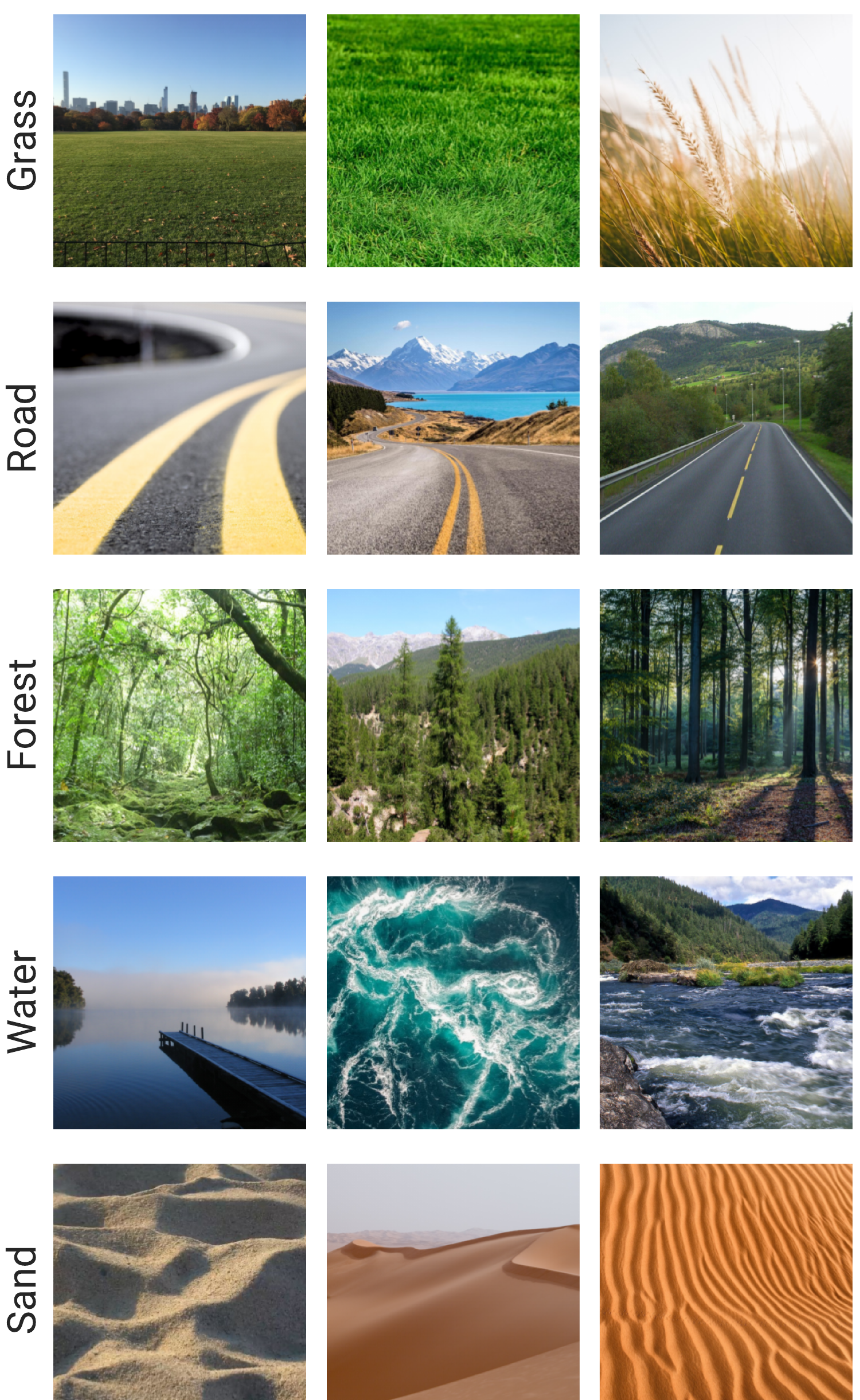}
    \end{subfigure}
    \begin{subfigure}[h]{0.45\textwidth}
    \includegraphics[trim={0 0 0 26cm}, clip, scale=0.5]{figures/backgrounds.pdf}
    \end{subfigure}

    \caption{Background templates used for generating images in D3S. See Sec. 3 of the main paper to see the algorithm for generation. Our algorithm is not sensitive to the exact templates; other images may also be used in the image guides.}
    \label{fig:backgrounds}
\end{figure*}

See \cref{fig:backgrounds}.

\begin{figure*}[h]
\centering
\includegraphics[scale=0.8]{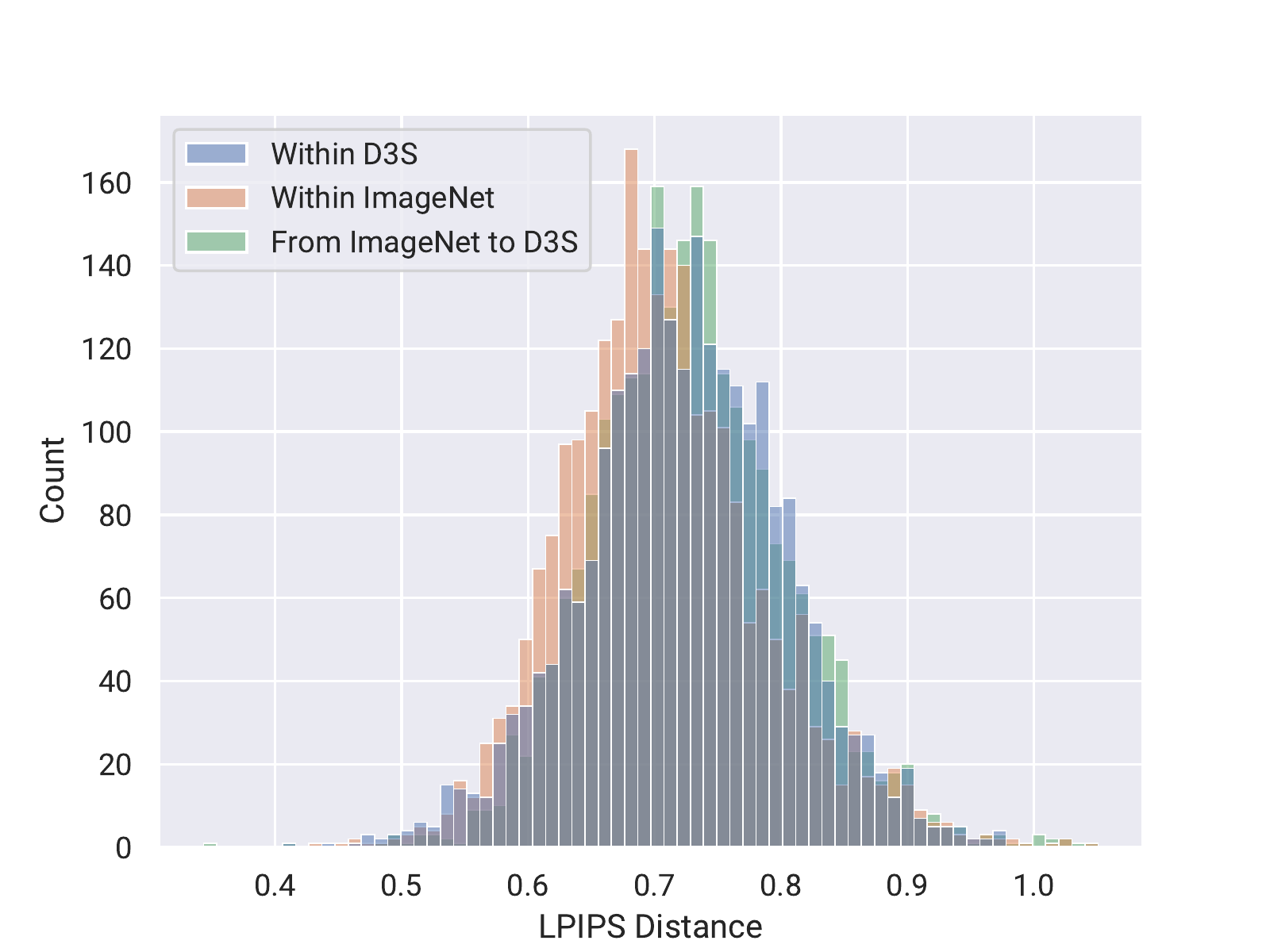}
\caption{Histogram of LPIPS distances~\cite{zhang2018unreasonable} distances between different pairs of images in D3S and ImageNet. Our dataset is perceptually as diverse as ImageNet.}
\label{fig:lpips}
\end{figure*}

\section{Diversity in D3S}
\Cref{fig:lpips} shows the histogram of LPIPS distances~\cite{zhang2018unreasonable} between 7500 pairs of images of the \textit{same} class where the images are:
\begin{enumerate}
    \item both drawn from D3S, shown in blue. This set has a mean LPIPS distance of 0.716.
    \item both drawn from ImageNet, shown in orange. This set has a mean LPIPS distance of 0.706.
    \item drawn one each from ImageNet and D3S. This is shown in green and the mean LPIPS distance of this set is 0.730.
\end{enumerate}
As we can see, our images are perceptually as diverse as those in ImageNet.

\section{Evaluating pretrained models on D3S}
We evaluate the following image classifiers on our D3S benchmark:
\begin{enumerate*}[label=(\roman*)]
\item Supervised CNNs (ResNet~\cite{he2016deep}, EfficientNet~\cite{tan2019efficientnet}, Inception-ResNet-v2~\cite{szegedy2017inception}, MobileNet~\cite{howard2018inverted, howard2019searching}, DenseNet~\cite{huang2017densely}, ResNeXt~\cite{xie2017aggregated}, ResNetv2~\cite{he2016identity})
\item Supervised vision transformers (ViT~\cite{dosovitskiy2021an}, DeiT~\cite{touvron2021training})
\item Adversarially robust models (Adversarial Inception networks \cite{kurakin2018adversarial}, Robust ResNets\cite{robustness})
\item Semi-supervised \& Semi-weakly supervised models (SSL, SWSL ResNets~\cite{zeki2019billion})
\item Self-supervised models (BEiT~\cite{bao2022beit}, DINO~\cite{caron2021emerging})
\item Clip zero shot classifiers ~\cite{radford2021learning}.
\end{enumerate*} \Cref{fig:all-gaps} shows the difference in top-1 accuracy of these models on the validation set of ImageNet and on our D3S benchmark. All models except the CLIP ones perform $\sim10-15\%$ worse on our dataset. The CLIP zero shot classifiers and StableDiffusion use the same text tokenizer. Additionally, the text prompts for zero shot classification are of the form ``a photo of a \{foreground\}'' which is a prefix of the text prompts we use for generating images in D3S. Therefore, when classifying D3S images with CLIP models, the CLIP image encoders extract embeddings which are closer to the above text prompts than in the case of images from ImageNet. This explains the increase in their classification accuracy on our dataset. Note that the hatched bars represent models that were trained with more data than that is in the train split of ImageNet.

\begin{figure*}[h]
\centering
\includegraphics[scale=0.55]{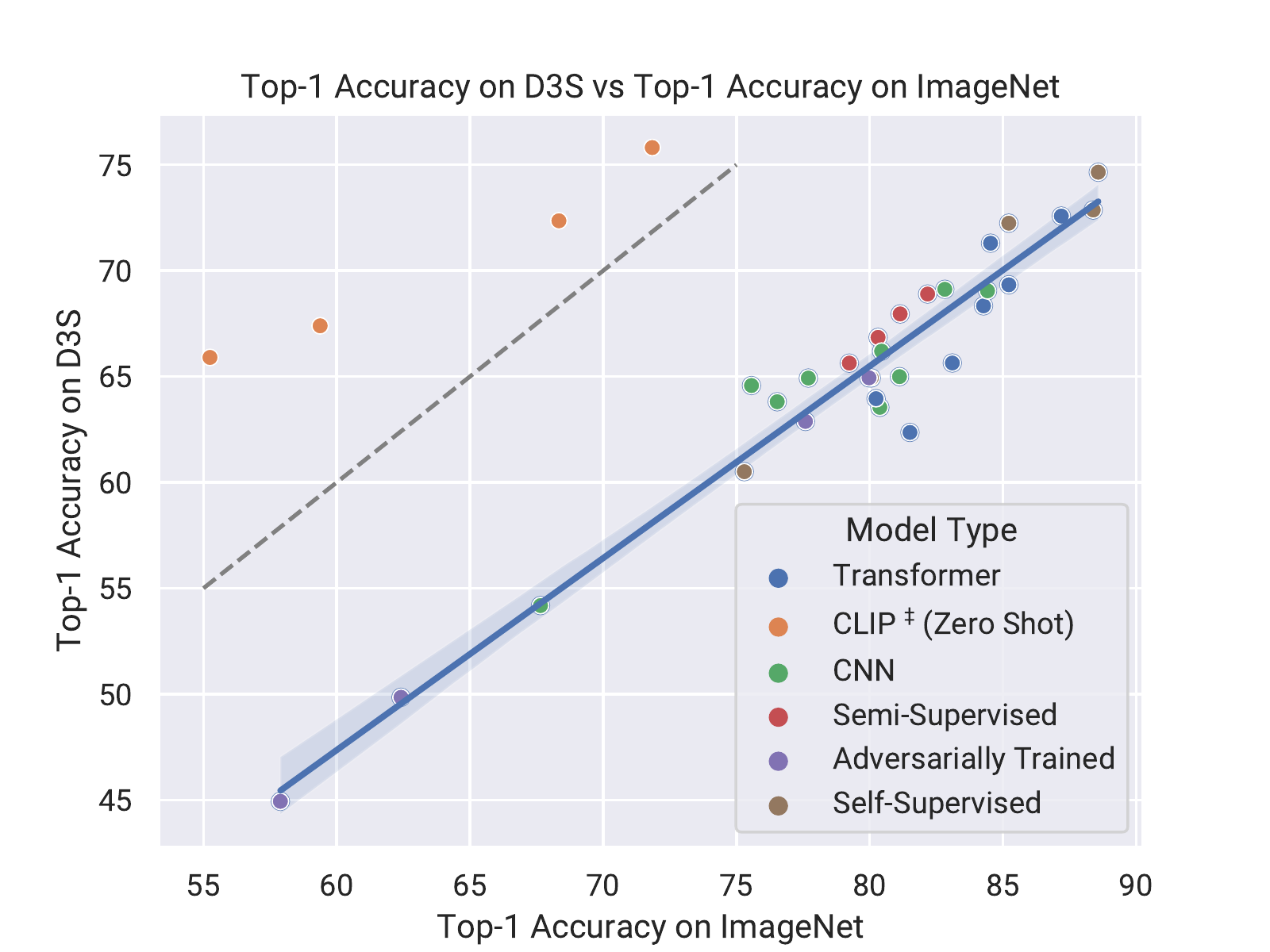}
\caption{Scatter plot of top-1 accuracies on D3S vs. top-1 accuracies on ImageNet for various models. The line of best fit is computed ignoring the CLIP models as they are outliers in this plot. $y=x$ is shown as a dashed line.}
\label{fig:scatter}
\end{figure*}

\Cref{fig:scatter} shows the scatter plot of top-1 accuracies on D3S vs. top-1 accuracies on ImageNet for the aforementioned models. Clearly, the CLIP zero-shot classifiers perform better than all other models. The line of best fit shown in the plot is computed based on all other models (excluding CLIP models). All semi-supervised models and most supervised CNNs perform relatively better than the trend, while most transformers and adversarially trained models perform worse.

\begin{figure*}[h]
\centering
\includegraphics[width=0.95\linewidth]{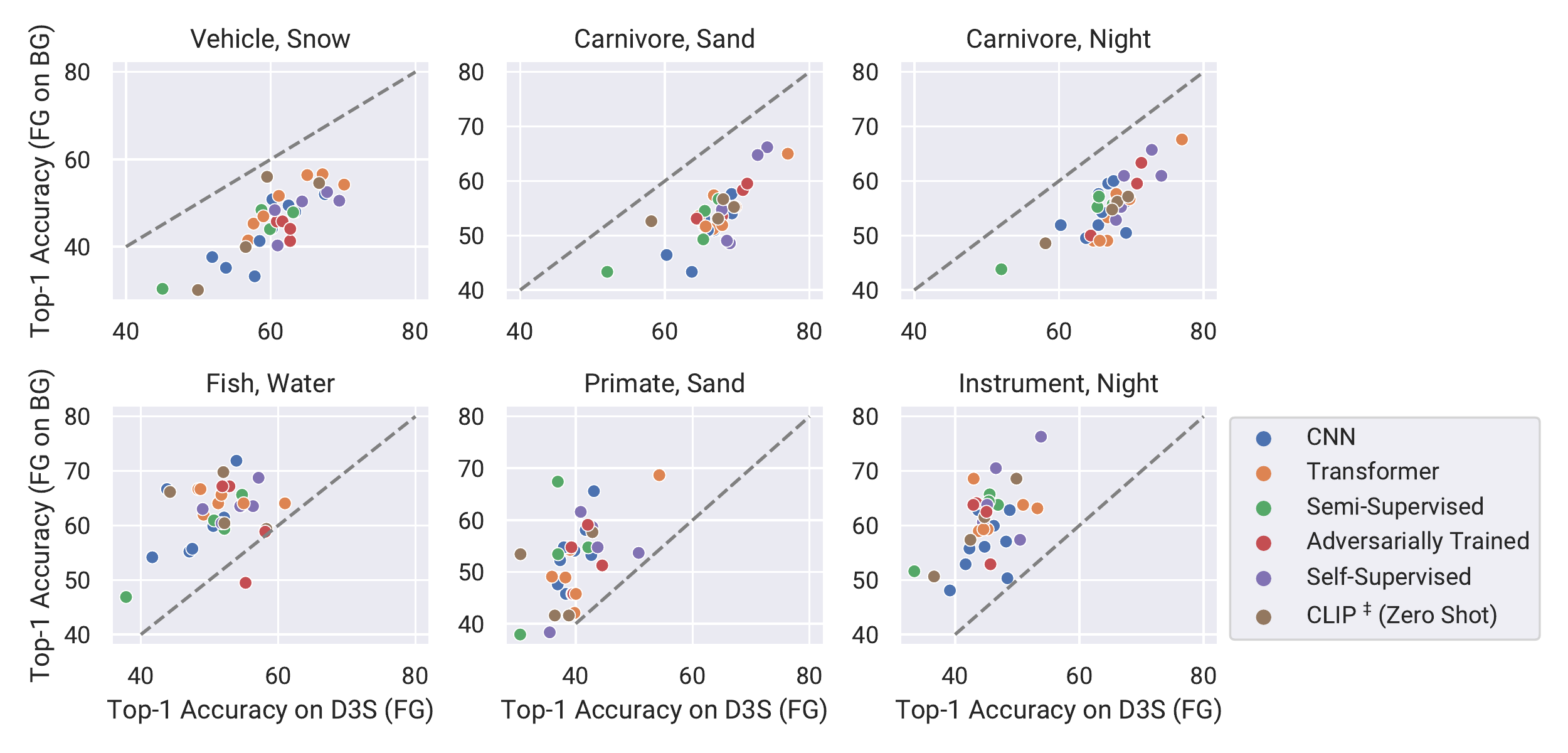}
\caption{Scatter plot of the foreground, background pairs for which the backgrounds cause the most adverse drop in foreground classification accuracy (top row) and those for which the backgrounds improve the accuracy the most (bottom row). See text for more details.}
\label{fig:robustness}
\end{figure*}

To identify the relative effect of backgrounds on various foregrounds, we first consolidate the 1000 ImageNet classes to 9 ImageNet-9 classes~\cite{xiao2021noise}. Then we plot the top-1 accuracy on images of a particular foreground, background pair as a function of the top-1 accuracy for that foreground in \cref{fig:robustness}. The top row shows the pairs for which the backgrounds have the most adverse affect (relatively) on the foreground classification accuracy: \begin{enumerate*}\item Vehicle in snow \item Carnivore in sand \item Carnivore in night \end{enumerate*}. Similarly, the bottom row shows the pairs for which the backgrounds improve the accuracy the most: \begin{enumerate*}
    \item Fish in water \item Primate in sand \item Instrument in night
\end{enumerate*}.
\begin{figure*}[h]
\centering
\includegraphics[width=0.95\linewidth]{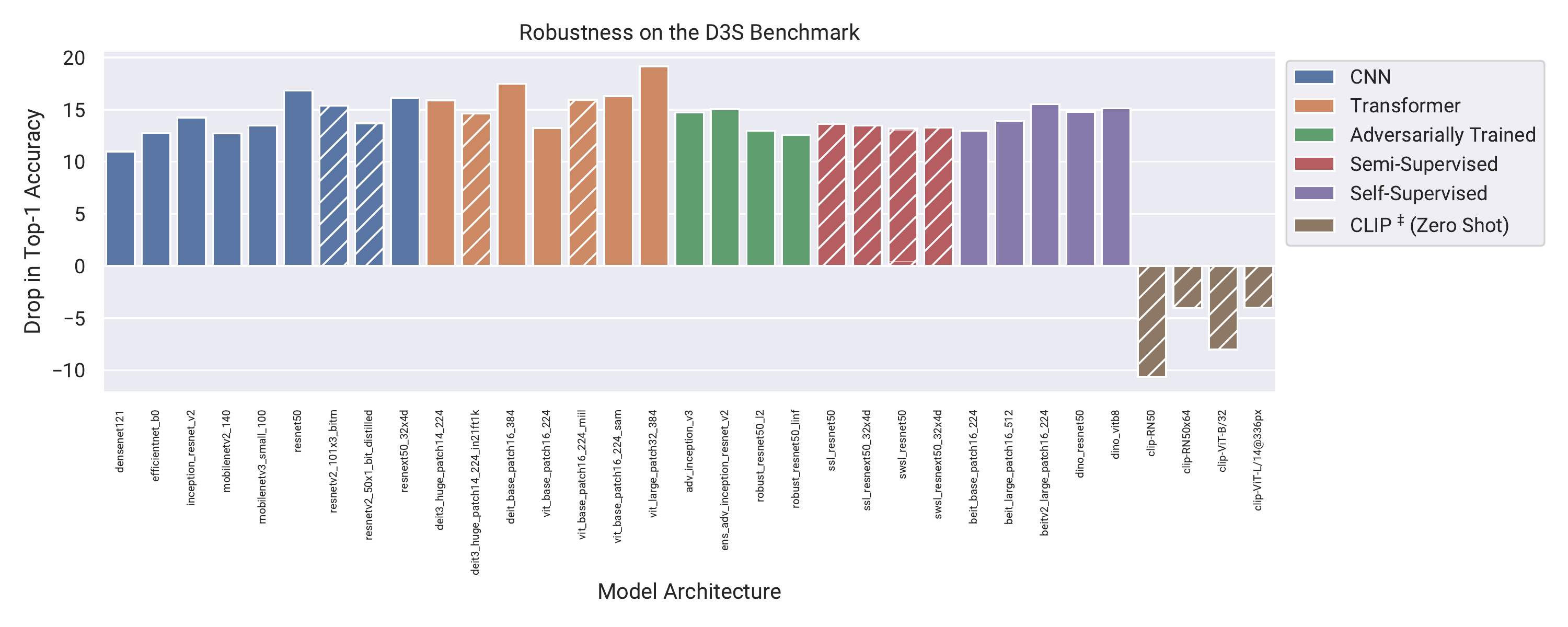}
\caption{Drop in top-1 accuracy of various models pretrained on ImageNet, when evaluated on our dataset D3S.}
\label{fig:gaps}
\end{figure*}
\section{Class pairs included in $D3S_{corr}$ and $D3S_{anticorr}$.}
\begin{figure*}[h]
    \centering
    \includegraphics[width=0.6\textwidth]{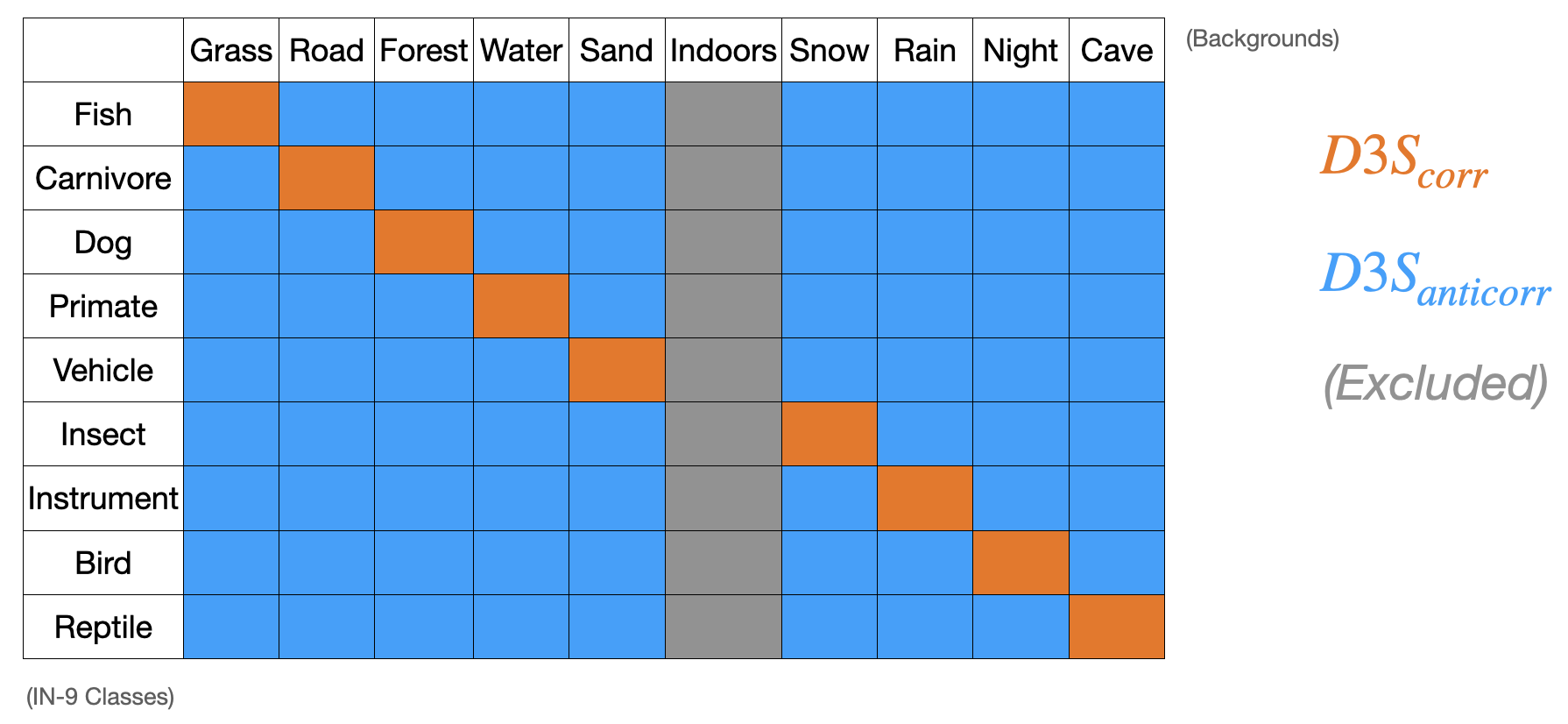}
    \caption{Foreground and Background class pairs included in $D3S_{corr}$ and $D3S_{anticorr}$.}
    \label{fig:corr_anticorr}
\end{figure*}

\Cref{fig:corr_anticorr} shows the (foreground, background) pairs we use to create the $D3S_{corr}$ and $D3S_{anticorr}$ partitions of our dataset. See Sec. 6 of the main paper for details on the experiment we run on these partitions.
\section{Experiments on Backgrounds Challenge}
\Cref{tab:backgrounds_challenge} shows the results of evaluating our Wasserstein Disentangled model on Backgrounds Challenge~\cite{xiao2021noise}. Specifically, the foregrounds of the images in this challenge are predicted using foreground features ($w_{fg}^TZ_{fg}$ in Sec. 5). ``Mixed-Same'' refers to images where foreground objects are pasted on random backgrounds of the same class. ``Mixed-Rand'' refers to image where they are pasted on random backgrounds of a random class. Adversarial images are constructed by pasting foregrounds on backgrounds are  chosen adversarially for each image. Our method performs worse than the baseline (ResNet50~\cite{he2016deep} pretrained on ImageNet). We hypothesize that this due to the simple ``cut and paste'' approach to generating these images. Such images are not natural looking unlike the images in our dataset and this might pose difficulty for our model in extracting the disentangled foreground features correctly.

\begin{table*}[h]
    \centering
    \begin{tabular}{|c|c|c|c|c|}
    \hline
    & Clean Images & ``Mixed-Same'' & ``Mixed-Rand'' & Adversarial\\
    \hline
        Wass. Disentanglement &91.38\%&78.44\%&69.68\%& 18.00\% \\
    \hline
        Baseline &  96.64\% & 90.42\% & 84.72\% & 22.62\%\\
    \hline
    \end{tabular}
    \caption{Results of Wasserstein Disentanglement model on Backgrounds Challenge~\cite{xiao2021noise}.}
    \label{tab:backgrounds_challenge}
\end{table*}
\section{Comparison of artificial distribution shift methods from prior work}
\begin{figure*}
    \centering
    \includegraphics[width=\textwidth]{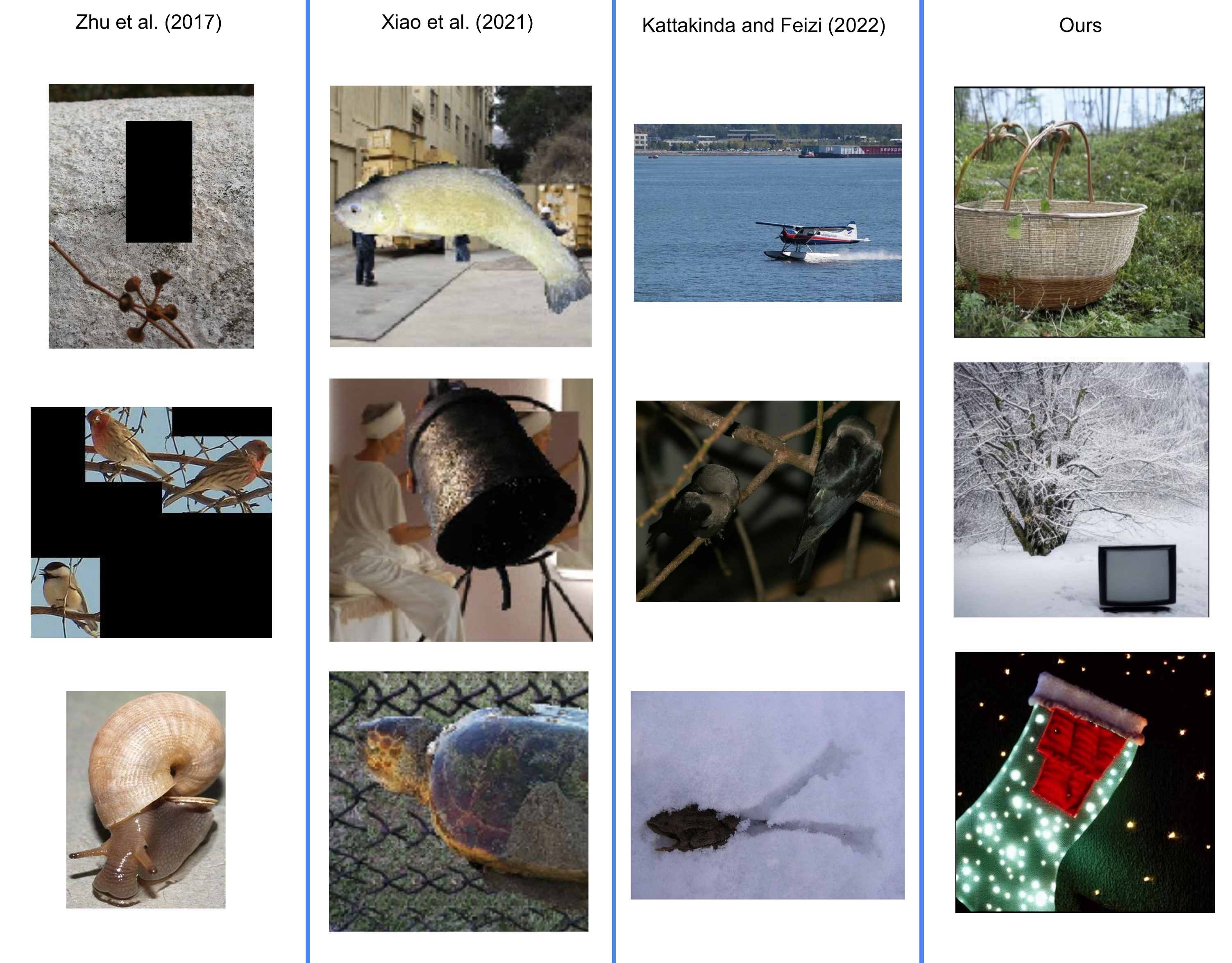}
    \caption{Comparison of methods to control foreground and background elements of images from literature.}
    \label{fig:compare_methods_lit}
\end{figure*}

Figure \ref{fig:compare_methods_lit} shows a comparison of the  ImageNet-like datasets with controlled background and foreground elements discussed in Section 2.1 in the main text. All images are sourced from the image examples in their respective papers, and may not necessarily be representative.
\section{Detailed foreground-background correlation experiment results}
Here we provide a version of Table 3 in the main text, with greater detail in the labeling of the datasets and models. See Table \ref{tab:accs_corr_expanded} below.
\begin{table*}[ht!]
    \centering
    \addtolength{\tabcolsep}{-0.2em}
    \begin{tabular}{|c!{\vrule width1.6pt}c|c|c|c|c|c|}
    \hline
        \multicolumn{1}{|r|}{Training Dataset} &  \multicolumn{5}{c|}{$D3S_{corr}$ (training set) }& \multirow{3}*{(Average)}  \\
        \cline{1-6}
         \multicolumn{1}{|r|}{Task} &  \multicolumn{3}{c|}{Foreground Prediction}& \multicolumn{2}{c|}{ Background Prediction}&  \\
        \cline{1-6}
         \multicolumn{1}{|r|}{Test Dataset}& $D3S_{corr}$ & $D3S_{anticorr}$& ImageNet & $D3S_{corr}$ & $D3S_{anticorr}$&  \\
         \Xhline{4\arrayrulewidth}
        \textbf{Wasserstein Disentanglement -}& \multirow{2}*{99.57\%}& \multirow{2}*{97.4\%}& \multirow{2}*{92.77\%}& \multirow{2}*{99.29\%}& \multirow{2}*{90.36\%}& \multirow{2}*{\textbf{95.88\%}}\\
       \textbf{Correct Features for Task (Ours)}  &&&&&&\\
         \hline
       Wasserstein Disentanglement - &\multirow{2}*{99.86\%}&\multirow{2}*{18.39\%}&\multirow{2}*{62.84\%}&\multirow{2}*{99.86\%}&\multirow{2}*{78.56\%}&\multirow{2}*{71.90\%}\\
        All Features &&&&&&\\
         \hline
       Baseline -&\multirow{2}*{99.72\%}&\multirow{2}*{91.71\%}&\multirow{2}*{94.55\%}&\multirow{2}*{99.72\%}&\multirow{2}*{6.485\%}&\multirow{2}*{78.44\%}\\
        ImageNet Pre-training &&&&&&\\
       \hline
    \end{tabular}
    \caption{Classification accuracy of linear classifiers trained that predict \textit{correlated} foreground or background label using our learned partitioned foreground and background features. Note that all foreground and ImageNet accuracies refer to ImageNet-9 classes.  ``Average'' is the average of the 5 experiments (not weighted to dataset size).  }
    \label{tab:accs_corr_expanded}
\end{table*}

\end{document}